\documentclass[journal]{IEEEtran}
\usepackage{graphicx}
\usepackage{booktabs}
\usepackage{times}
\usepackage{soul}
\usepackage{color}
\usepackage{url}
\usepackage{amsfonts,amssymb}
\usepackage[hidelinks]{hyperref}
\usepackage[utf8]{inputenc}
\usepackage[small]{caption}
\usepackage{graphicx}
\usepackage{amsmath}
\usepackage{amsthm}
\usepackage{booktabs}
\usepackage{algorithm}
\usepackage{xcolor} 
\usepackage{algorithmic}
\usepackage[switch]{lineno}
\usepackage{booktabs}
 \usepackage{multirow}
 \usepackage{pifont}
 \usepackage{bm}
 \usepackage{adjustbox}
 \usepackage{makecell}
\usepackage{array}  
\usepackage{tabularx} 
\usepackage[numbers,sort&compress]{natbib}

\ifCLASSINFOpdf
\else
\fi
\begin{document}
%
\title{LogiCode: an LLM-Driven Framework for Logical Anomaly Detection}


%
%
%


\author{Yiheng~Zhang,
        Yunkang~Cao,~\IEEEmembership{Graduate Student Member,~IEEE,}
        Xiaohao~Xu,
        Weiming Shen\IEEEauthorrefmark{1},~\IEEEmembership{Fellow,~IEEE,}
\thanks{Weiming Shen\IEEEauthorrefmark{1} (wshen@ieee.com) is the corresponding author.}
\thanks{Manuscript received April xx, 2023;}}

\markboth{Submitted to IEEE TRANSACTIONS ON AUTOMATION SCIENCE AND ENGINEERING}%
{Shell \MakeLowercase{\textit{et al.}}: Bare Demo of IEEEtran.cls for IEEE Journals}
\maketitle

\IEEEpeerreviewmaketitle
\begin{abstract}
This paper presents LogiCode, a novel framework that leverages Large Language Models (LLMs) for identifying logical anomalies in industrial settings, moving beyond traditional focus on structural inconsistencies. By harnessing LLMs for logical reasoning, LogiCode autonomously generates Python codes to pinpoint anomalies such as incorrect component quantities or missing elements, marking a significant leap forward in anomaly detection technologies. A custom dataset “LOCO-Annotations” and a benchmark “LogiBench” are introduced to evaluate the LogiCode’s performance across various metrics including binary classification accuracy, code generation success rate, and precision in reasoning. Findings demonstrate LogiCode’s enhanced interpretability, significantly improving the accuracy of logical anomaly detection and offering detailed explanations for identified anomalies. This represents a notable shift towards more intelligent, LLM-driven approaches in industrial anomaly detection, promising substantial impacts on industry-specific applications.
\end{abstract}
\renewcommand{\abstractname}{Note to Practitioners}
\begin{abstract}
This work introduces LogiCode, an innovative system leveraging Large Language Models (LLMs) for logical anomaly detection in industrial settings, shifting the paradigm from traditional visual inspection methods. LogiCode autonomously generates Python codes for logical anomaly detection, enhancing interpretability and accuracy. Our novel approach, validated through the “LOCO-Annotations” dataset and LogiBench benchmark, demonstrates superior performance in identifying logical anomalies, a challenge often encountered in complex industrial components like assembly and packaging. LogiCode provides a significant advancement in addressing the nuanced requirements of detecting logical anomalies, offering a robust and interpretable solution to practitioners seeking to enhance quality control and reduce manual inspection efforts.
\end{abstract}

\begin{IEEEkeywords}
Logical Anomaly Detection, Large Language Models, Industrial Anomaly Detection, Dataset Annotation.
\end{IEEEkeywords}
\section{Introduction}

\begin{figure}[t!]
\centering\includegraphics[width=\linewidth]{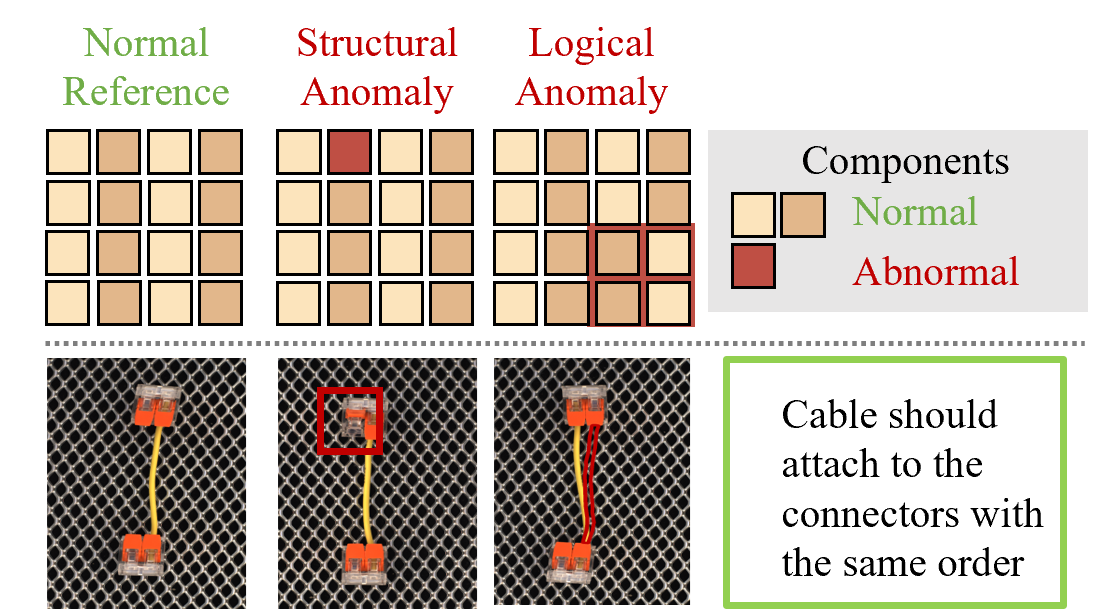}
\caption{Comparison between the structural anomaly and the logical anomaly. (\textbf{Top}) Toy example. (\textbf{Bottom}) Examples from the MVTec LOCO AD dataset~\cite{Bergmann2022Beyond}. In toy example, the yellow squares represent normal components, while the red squares signify structural deviations from the norm. Notably, even though the individual components in logical anomaly are normal, they collectively defy logical constraints. For the LOCO examples, structural anomaly is exemplified by damaged connector, while logical anomaly is indicated by misplaced cable connections.}
\label{fig:tasksetting}
\end{figure}

\IEEEPARstart{A}{nomaly} detection in industrial scenarios is pivotal for ensuring the quality and reliability of products ~\cite{cao2024survey},~\cite{dong2021defect},~\cite{Yao2023DualAttention}. Traditionally, this domain has been dominated by methods focusing on structural defects, such as dents or scratches~\cite{bergmann2019mvtec}, detectable through vision-based techniques. However, these conventional methods often fall short when dealing with logical anomalies~\cite{Bergmann2022Beyond} – errors in high-level semantic logic, such as incorrect orders in combinations of normal components, as illustrated in Fig.~\ref{fig:tasksetting}. These logical anomalies widely exist in industrial settings, like assembly and packaging. Given the potential for severe functional consequences associated with logical anomalies, specific logical anomaly detection methods have been developed.

\begin{figure}[t!]
\centering\includegraphics[keepaspectratio]{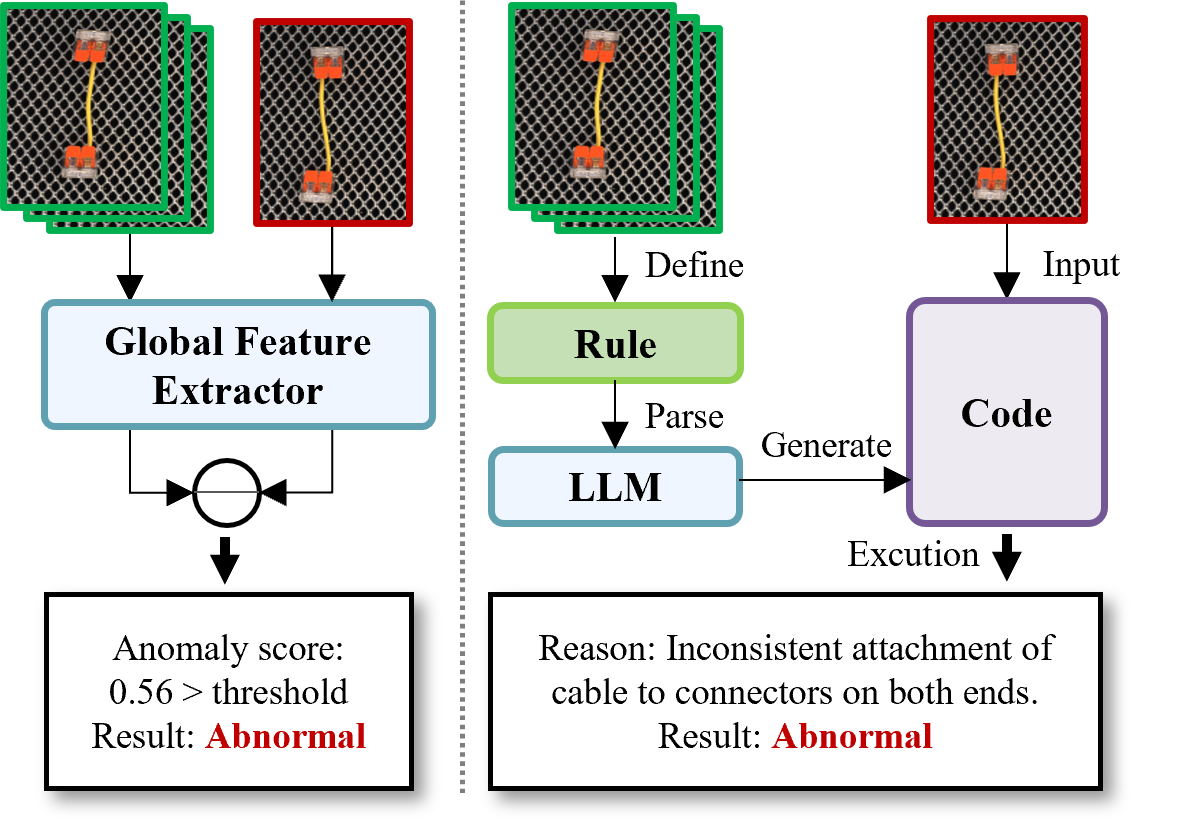}
\caption{Comparison between the existing method (\textbf{left}) and our LogiCode framework (\textbf{right}). The existing method do not accurately capture the underlying logical relationships  and lack explanation for logical anomalies. In contrast, the proposed method starts from the perspective of logical constraints and gives the result with reason.}
\label{fig:GCADcompare}
\end{figure}

Existing logical anomaly detection methods, such as GCAD~\cite{Bergmann2022Beyond} and its successors~\cite{Yang2023SLSG},~\cite{Yao2023Learning},~\cite{Batzner2024Efficientad}, have made their initial progresses, as shown in Fig.~\ref{fig:GCADcompare}. These methods leverage both local and global network branches to detect structural and logical inconsistencies, with a particular emphasis on global consistency representations. They assume that global consistency representations can reflect the overall logical constraint of an image. However, such representations often fail to equate to the deep logical relationships necessary for identifying subtle logical anomalies—like an incorrect sequence of correctly quantified components or components that match in type but not in specification. The inherent challenge lies in detecting these high-level semantic inconsistencies, which requires an understanding of real logical reasoning and relationships beyond the capabilities of purely vision-based systems. Furthermore, these methods often present results for logical anomalies through anomaly scores(see Fig.~\ref{fig:GCADcompare}).  However, unlike structural anomaly detection, logical anomaly detection typically demands clearer explanations while the lack of explicit reasoning can lead to confusions. This highlights the need for a method that not only detects but also articulately explains logical anomalies.

Recent advances in LLMs~\cite{gupta2023visual},~\cite{lai2023lisa},~\cite{you2023idealgpt}, offer new insights and solutions for this challenge. LLMs excel in understanding and processing complex logical relationships and reasoning~\cite{Shen2023Hugginggpt},~\cite{Suris2023Vipergpt}, making them particularly suited for detecting and analyzing these logical anomalies. Unlike traditional methods, LLMs can interpret and reason about high-level semantic logic, providing an innovative approach to anomaly detection.

Hence, this study proposes a LogiCode framework, utilizing the advanced capabilities of LLMs, representing a paradigm shift in logical anomaly detection. LogiCode, utilizing expert knowledge, extracts logical relationships from normal images and defines them as a set of logical rules. These rules are then parsed by the LLM, which automatically generates executable codes to examine the consistency between testing images and normal rules. Furthermore, the framework is designed to provide reasons for these anomalies, enhancing the interpretability of its findings, as shown in Fig.~\ref{fig:GCADcompare}.

In this framework, the LLM's process of analyzing visual information and detecting logical anomalies is akin to the interplay between human perception and cognitive reasoning. LogiCode embodies this synergy, offering a more natural and intuitive approach to anomaly detection that mirrors human expertise. The framework transcends the capabilities of traditional models by providing enhanced interpretability, direct logic interpretation, and adaptability to the complex nuances of industrial quality control.

The key contributions of this study include:

\begin{itemize}
    \item It proposes a new approach in logical anomaly detection with LogiCode, a framework that leverages the logical reasoning capabilities of LLMs to comprehend and interpret logical relationships. By generating executable Python codes, LogiCode emulates human-like reasoning, enabling precise detection and explanation of logical anomalies in industrial scenarios. This innovative approach marks a significant advancement in the field of logical anomaly detection.
    \item It introduces LogiBench, a benchmarking tool that integrates LLM automatic evaluation with expert manual analyses to evaluate the efficacy of LogiCode. LogiBench serves as a comprehensive measure for assessing the accuracy and effectiveness of logical anomaly detection systems.
    \item It also contributes to the field by introducing the LOCO-Annotations dataset. It is meticulously designed to deeply analyze logical anomaly detection, encompassing a wide range of scenarios and providing a rich resource for both automatic and manual evaluations. To facilitate further research in logical anomaly detection, we have publicly released the LOCO-Annotations dataset at https://github.com/22strongestme/LOCO-Annotations.
\end{itemize}

Through these innovations, LogiCode not only addresses the existing gaps in logical anomaly detection but also enhances interpretability and adaptability in this field. Moreover, it sets a new benchmark in the application of LLMs for logical anomaly detection as a precedent for future exploration. 

The reminder of this paper is organized as follows. Section I provides a comprehensive overview of the proposed framework and its position within the context of current research. Section II reviews some related work and establishes the foundation for the LogiCode framework. Section III details the methodological approach of LogiCode, followed by Section IV, which presents an in-depth analysis of experimental results and performance evaluations. Finally, Section V discusses the implications of our findings and explores potential directions for future research in logical anomaly detection using LLMs.

\section{Related Works}
This section delves into advancements in anomaly detection, including structural and logical approaches, as well as studies utilizing LLMs for automation tasks.

\subsection{Structural Anomaly Detection}
Existing methods that excel in detecting low-level structural anomalies can be categorized into two main types: reconstruction-based and feature-embedding similarity-based approaches.

Reconstruction-based methods, such as Autoencoders~\cite{Hinton2006Reducing} (AEs) and Generative Adversarial Networks~\cite{Akcay2019Ganomaly} (GANs), attempt to reconstruct input images through a lower-dimensional bottleneck. In this context, Yang et al.~\cite{Yang2019Multiscale} proposed a multiscale feature-clustering-based fully convolutional autoencoder that significantly improves the speed and accuracy of visual inspection for texture surface defects. Further advancing this category, innovations like RIAD~\cite{Zavrtanik2021Reconstruction} reformulates reconstruction as an image inpainting problem. While these methods are efficient in capturing image contexts~\cite{Bergmann2018Improving}, they often yield blurry reconstructions, leading to increased false positives.

On the other hand, feature-embedding based methods, such as RD4AD~\cite{Deng2022Anomaly}, Padim~\cite{Defard2021Padim}, PatchSVDD~\cite{Yi2020PatchSVDD}, SPADE~\cite{Cohen2020SubImage} and PatchCore~\cite{Roth2022Towards}, utilize deep neural network-extracted vectors representing the entire image. Cao et al.~\cite{Cao2022Informative} explored informative knowledge distillation within image anomaly segmentation to enhance model learning from complex data distributions. Moreover, Jiang et al.~\cite{Jiang2023Masked} introduced a masked reverse knowledge distillation approach that leverages both global and local information for enhancing image anomaly detection. These methods, relying on the distance between the embedding vectors of test images and normal reference vectors from the training dataset~\cite{yao2022feature},~\cite{Bergmann2020Uninformed}, demonstrate superior performance compared to reconstruction-based approaches~\cite{Cao2023Collaborative}. 

Both reconstruction-based and feature-embedding based approaches have shown great result, on datasets for structural anomalies like MVTec AD~\cite{bergmann2019mvtec}. However, these methods, constrained by their limited receptive fields, struggle to detect anomalies beyond receptive fields and fail to distinguish violations of logical constraints. This is particularly true in LOCO datasets, where the detection of logical anomalies is crucial. This highlights a significant gap in current anomaly detection methodologies, underlining the need for more adaptable and comprehensive systems that can effectively detect logical anomalies.

\subsection{Logical Anomaly Detection}
In addressing high-level semantic logic anomalies, several methods have emerged in recent research. MVTec LOCO~\cite{Bergmann2022Beyond}, as a milestone, is specialized for studying and resolving advanced semantic logic anomalies. Alongside, GCAD~\cite{Bergmann2022Beyond} was proposed to solve logical anomaly detection. GCAD includes  a local network branch and a global network branch. The local branch is designed to detect novel local structures in images, while the global branch learns a global consistency representation through a bottleneck structure, targeting violations in long-distance dependencies. This structure allows for the detection of both structural and logical anomalies. Subsequent methods, focusing on logic detection, often continue to employ the strategy of GCAD’s global network branch, which is based on vision-based global-local correspondences. For example, EfficientAD~\cite{Batzner2024Efficientad} utilizes an autoencoder-student network pair to detect anomalies that violate global semantic constraints. Similarly, GLCF~\cite{Yao2023Learning} achieves this through a combination of local and global network branches; the local branch focuses on structural anomaly detection, while the global branch captures logical anomalies through semantic bottlenecks. However, global feature extraction may not always discern subtle logical anomalies, like correct component quantities arranged in the wrong order, or components of the right type but incorrect specifications. Therefore, employing the logical reasoning capabilities of LLMs offers a more universal appraoch for resolving logical anomalies.

\subsection{LLMs in Automation Tasks}
The advance of LLMs marks a significant milestone in the field of automation, especially in industrial applications. These models have transcended traditional boundaries, offering novel solutions in code generation~\cite{Suris2023Vipergpt}, visual question answering (VQA)~\cite{Shen2023Hugginggpt},~\cite{johnson2017clevr}, and anomaly detection~\cite{Gu2023AnomalyGPT}.

ViperGPT ~\cite{Suris2023Vipergpt}, a frontrunner in this domain, exemplifies the integration of LLMs in VQA, using code-generation to resolve visual queries more effectively than traditional methods. This represents a significant advancement over previous code generation techniques, providing more context-aware and adaptable solutions. The industrial application of LLMs extends to defect detection, anomaly detection, and quality control. Zhang et al.~\cite{Zhang2023Exploring} demonstrate its capability in zero-shot anomaly detection, applying it to complex VQA tasks. Wang et al.~\cite{Wang2023Industrial} introduces Industrial-GPT, tailored for intelligent manufacturing and excelling in tasks like fault diagnosis. Furthermore, Cao et al.~\cite{Cao2023Towards} illustrates LLM's effectiveness for anomaly detection in multi-modal domains, while also acknowledging some limitations in intricate scenarios. AnomalyGPT~\cite{Gu2023AnomalyGPT} also emerges as an innovative industrial anomaly detection model, aligning images with textual descriptions to improve anomaly detection.

By leveraging the advancements in LLMs for logical reasoning and code generation, and addressing the limitations of existing logical anomaly detection methods, LogiCode represents a significant evolution in tackling high-level semantic logic anomalies in industrial settings.

\begin{figure*}[ht!]
\centering\includegraphics[keepaspectratio]{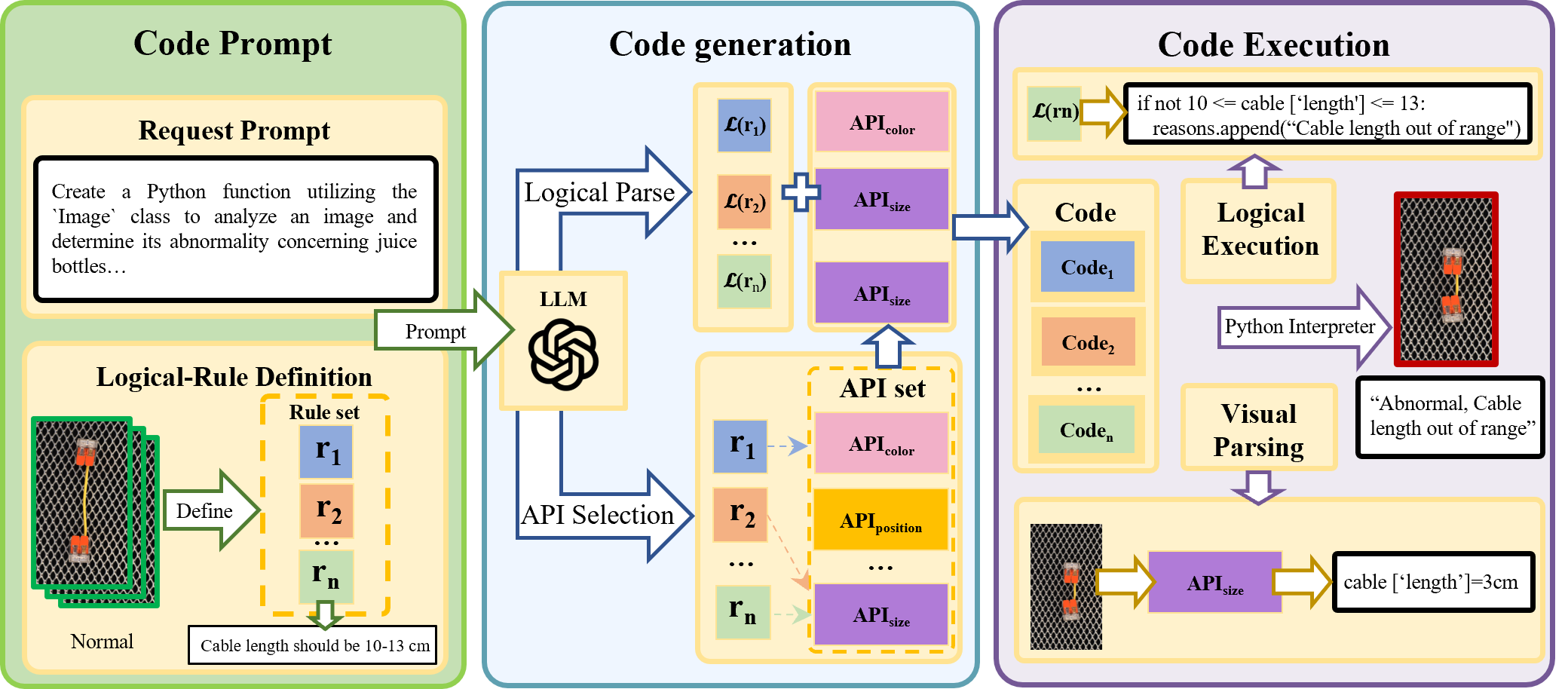}
\caption{\textbf{Overview of LogiCode framework} LogiCode is an LLM-empowered logical anomaly detection framework. It comprises three main modules: \textit{Code Prompt}, \textit{Code Generation}, and \textit{Code Execution}. \textbf{a}) \textit{\textbf{Code Prompt}} module formulates user-defined tasks and logical rules by analyzing examples of normal and abnormal images. \textbf{b}) \textit{\textbf{Code Generation}} module utilizes an LLM to parse these rules into executable Python codes, selecting appropriate APIs for image analyses. \textbf{c}) \textit{\textbf{Code Execution}} module applies logical and visual parsing to detect and report anomalies. This comprehensive process not only detects anomalies but also provides rule-based explanations for them, mimicking human problem-solving and decision-making abilities in industrial settings.}
\label{fig:framework}
\end{figure*}
\section{Proposed LogiCode Framework}
The LogiCode framework is an LLM-empowered framework developed for identifying logical anomalies in images. It is systematically introduced in the following subsections.

\subsection{Framework Overview}
The LogiCode framework is structured into three interconnected main modules—Code Prompt, Code Generation, and Code Execution, as shown in Fig.~\ref{fig:framework}. Drawing inspiration from human cognitive processes, the framework adeptly combines observation and logical reasoning in its modules. The Code Prompt module initiates the process of identifying problems and formulating logical rules, akin to human problem recognition and solution crafting. Code Generation, powered by an LLM, processes these rules to create logical decision codes and selects corresponding visual APIs, reflecting human cognitive synthesis for decision-making. Code Execution module integrates logical evaluation with visual parsing, analogous to human reasoning and information interpretation, focusing on essential visual details. This module’s output offers a detailed analysis, identifying anomalies with rule-based explanations, showcasing the framework’s human-like problem-solving approach. The collective functionality of these modules highlights how the LogiCode framework excels in logical anomaly detection and interpretability, both essential for industrial applications.

\subsection{Code Prompt}
Code Prompt serves as the initial stage of the logical anomaly detection process. This module combines the functionalities of “Request Prompt” and “Logical-Rule Definition” to define the analyses task and establish the corresponding logical rules:

\textbf{Request Prompt:} Within “Code Prompt”, the “Request Prompt” sub-module is meticulously designed to formulate the user’s task definition into a structured prompt for the utilized LLM. This module involves several components, each playing a critical role in transforming complex tasks into structured LLM prompts:

\textit{Task Interpretation:} Request Prompt starts with a clear interpretation of the task, as in “Create a Python function utilizing the Image class to comprehensively analyze an image and determine its abnormality concerning its cable and connectors layout,” ensuring task needs are precisely articulated in LLM-compatible formats.

\textit{Function Structuring:} Simultaneously, Request Prompt organizes the prompt to include all critical aspects of the function, such as inputs and expected outputs, ensuring that instructions like “the function should accept the image path and return two results: a boolean and a string listing all the reasons for abnormality,” are clearly communicated to the LLM.

\textit{Knowledge Integration:} Domain-specific knowledge is then woven into the Request Prompt, guiding the system to “Employ basic Python features for logical operations and mathematical calculations,” thus grounding the task in the relevant technical context.

\textit{Prompt Engineering:} Inspired by the latest research in prompt engineering~\cite{white2023prompt},~\cite{ye2023prompt},~\cite{kojima2022large}, this component incorporates strategies like outlining sub-tasks and illustrating relationships, like “Initially, outline the sub-tasks required for the analyses. Illustrate the relationships between these sub-tasks through a detailed step-by-step breakdown.” This approach guides the LLM in understanding the structure and logic of the task, enhancing the accuracy and efficiency of the code generation process.

\textbf{Logical-Rule Definition:} Within “Code Prompt”, the “Logical-Rule Definition” is a critical sub-module that operates in conjunction with the “Request Prompt”. It is focused on defining logical rules with expert knowledge from observations of  normal example images.

Each set of rules is scenario-specific and can be adapted to fit a wide range of contexts, making the framework versatile and applicable across different industries. The LogiCode framework, through this sub-module, showcases its adaptability by allowing these rules to be modified or expanded based on the requirements of different industrial scenarios, thereby maintaining the relevance and efficacy of the anomaly detection process.

\begin{table*}[ht!]
\centering
\caption{Illustration for API set functions}
\centering\includegraphics[keepaspectratio]{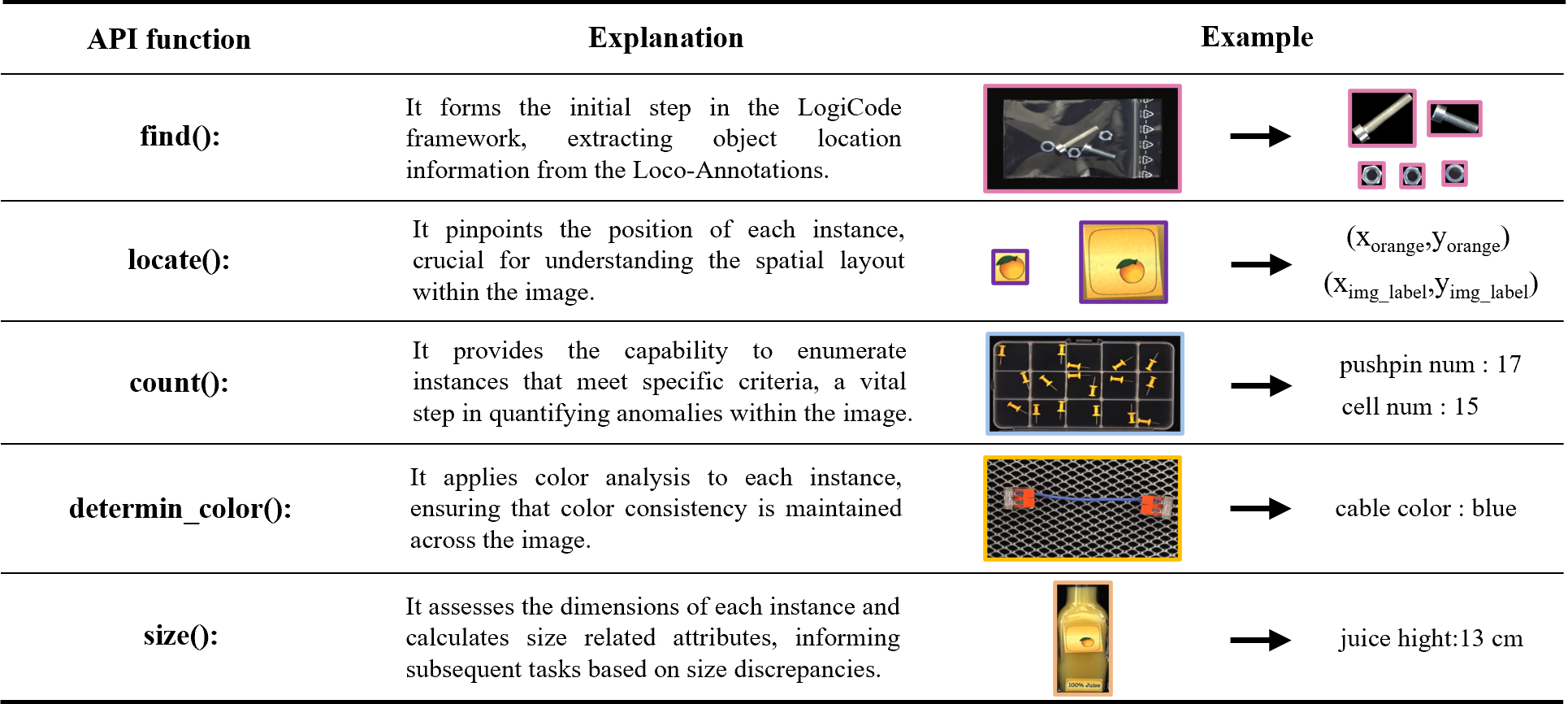}
\label{tab:apiset}
\end{table*}

\subsection{Code Generation}
Following Code Prompt, Code Generation in the LogiCode framework serves as a crucial bridge between conceptual rules and executable codes. This module harnesses the power of the LLM to transform structured request prompts and logical rules into Python codes. The process involves several key components:

\textbf{Logical Parsing:} Once the LLM receives the structured prompts, it employs its extensive knowledge base to parse complex logical conditions into executable Python codes. For example, a rule concerning the length of cable is interpreted by the LLM as:
\begin{equation}
L(r_{\text{length}}) : \text{if } L_{\text{cable}} \notin [L_{\text{min}}, L_{\text{max}}]
\end{equation}

$L_{\text{cable}}$ is the measured cable length, and $[L_{\text{min}}, L_{\text{max}}]$ is the acceptable length range for the cable. Each rule $r$ from the set $R$ is parsed into a logical expression $L(r)$ that the Python interpreter can execute.

\textbf{API Selection:} The designed Image class provides a rich API set that the LLM selects from to match the logical rules with the functional capabilities needed for image analyses, as shown in Table~\ref{tab:apiset}. This selection is crucial as it dictates the precision and efficiency of the subsequent image processing. The LLM matches each logical rule $r$ from the set $R$ with an API function $\mathcal{A}(r)$ from the Image class, ensuring that each aspect of the image analyses is addressed.

Each method $\mathcal{A}(r)$ in the API set is chosen for its ability to provide the necessary data for the logical rules to operate on. For example, the size() method, used in adherence to a rule regarding object size, is synthesized into the code as follows:
\begin{equation}
\mathcal{A}(r_{\text{length}}) : API_{\text{size}}(\text{image}) \rightarrow L_{\text{cable}} 
\end{equation}
This expression normalizes how the size() method is selected and utilized to extract the length of cable, denoted by $L_{\text{cable}}$, from the given image. It encapsulates the action of the LLM in choosing the size() method from the Image class when the rule $r_{\text{length}}$ is concerned with verifying the length of the cable within acceptable limits.

\textbf{Code Synthesis:} After logical parsing and API selection, the LLM synthesizes the information into a cohesive block of Python code. This synthesis combines the logic and API calls into a sequence of executable instructions:
\begin{equation}
     c = \bigcup_{r \in R} \{\mathcal{L}(r), \mathcal{A}(r)\} 
\end{equation}

The SynthesizeCode component illustrates the LLM's action in compiling the parsed rules and selected APIs into a final code segment ready for execution.

Code Generation thus serves as a crucial bridge, translating the structured input from the Code Prompt into Python codes that are ready for execution. It uses LLM's inferential power and the APIs' feature extraction capabilities to perform detailed image analyses and accurately identify logical anomalies.

\begin{figure*}[h]
\centering\includegraphics[width=1.\linewidth]{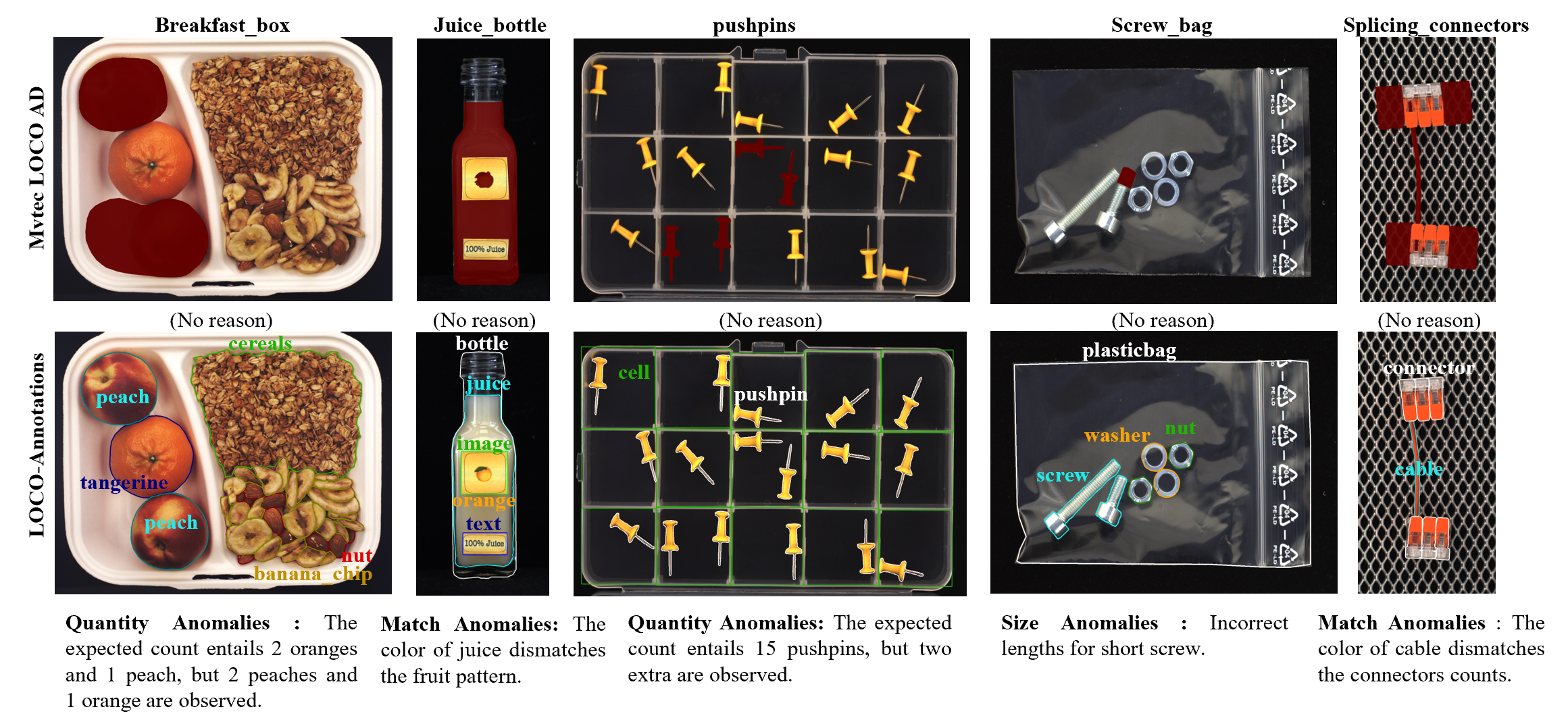}
\caption{The diagram contrasts the MVTec LOCO dataset (\textbf{Top}) with the LOCO-Annotations (\textbf{Bottom}). Unlike MVTec’s broad focus, the LOCO-Annotations dataset zeroes in on logical anomalies, offering an enriched dataset with detailed explanations for each anomaly. This provides LLMs with the context and specificity required for a nuanced understanding of anomalies, thereby facilitating a more sophisticated and targeted development in industrial quality control systems.}
\label{fig:LOCO-Annotations}
\end{figure*}

\subsection{Code Execution}

Code Execution is the operational core of the LogiCode framework where the Python codes synthesized by Code Generation are brought to execution. This module interprets and runs the codes on the given image data, applying the logical rules and utilizing the APIs to perform a detailed analysis of the imagery. It consists of two critical sub-modules, each responsible for different aspects of the execution process:

\textbf{Logical Execution:} This sub-module is responsible for the logical assessment of each image based on the generated codes. It executes the conditional statements and loops derived from the logical parsing process, directly applying them to the image data. For instance, the rule concerning the cable length is translated into an executable code that is conducted as depicted in the Code Execution section of Fig.~\ref{fig:framework}. The logical execution is facilitated by the Python interpreter, which assesses the conditions set by each rule against the actual image data.

\textbf{Visual Parsing:} Integrated within Code Execution, visual parsing is crucial for the extraction of visual features from the image that are necessary for logical anomaly detection. This sub-module uses the selected APIs to determine the size, position, color, and other relevant attributes of the image components. The process of measuring the cable’s length in the image, for instance, is exemplified in the orange section of the figure. The outcome of size is subsequently appraised by the predefined logical rules to ascertain its congruence with the acceptable size range. 

Then the execution process integrates the outputs from logical execution and visual parsing, cross-referencing them to identify any anomalies. The integration ensures that all the visual features extracted are checked against the logical conditions, providing a comprehensive and accurate assessment.

The final output from Code Execution includes a detailed analysis report, highlighting any detected anomalies with their corresponding reasons. The report is designed to be both thorough and interpretable, allowing for quick identification of issues and facilitating decision-making processes.

By combining logical reasoning with visual data interpretation, Code Execution encapsulates the functionality required to identify logical anomalies accurately, reflecting a significant advancement in the application of LLMs to industrial anomaly detection.

\section{Evaluations and Discussions}
This section evaluates the proposed LogiCode framework’s logical anomaly detection in industrial contexts using a specialized dataset, the LOCO-Annotations. It includes thorough benchmarking with LogiBench to test classification accuracy, code generation, and reasoning about anomalies. These tests aim to validate the LogiCode’s capabilities for industrial applications. Then the effectiveness and impact of LogiCode are discussed, highlighting its technological advancements and potential for application.

\subsection{LOCO-Annotations Dataset}
This subsection introduces the LOCO-Annotations dataset, a specialized extension of the MVTec LOCO dataset, aimed at filling a critical gap in the realm of logical anomaly detection in industrial scenarios. While the original MVTec LOCO dataset offers a robust foundation with its mix of structural and logical anomalies in industrial images, it primarily adheres to unsupervised anomaly detection, with a significant emphasis on detecting regions indicative of logical anomalies. Yet, the practical application of anomaly detection often necessitates an in-depth understanding of the underlying causes of these logical anomalies. In response to this, LOCO-Annotations shifts its focus from mere anomaly region detection to a thorough analysis of the underlying reasons for logical anomalies. This pivot aligns with the nuanced requirements of detecting and understanding logical anomalies, thereby better addressing the complex quality control demands in industrial environments.

\begin{figure}[h!]
\centering\includegraphics[width=1.\linewidth]{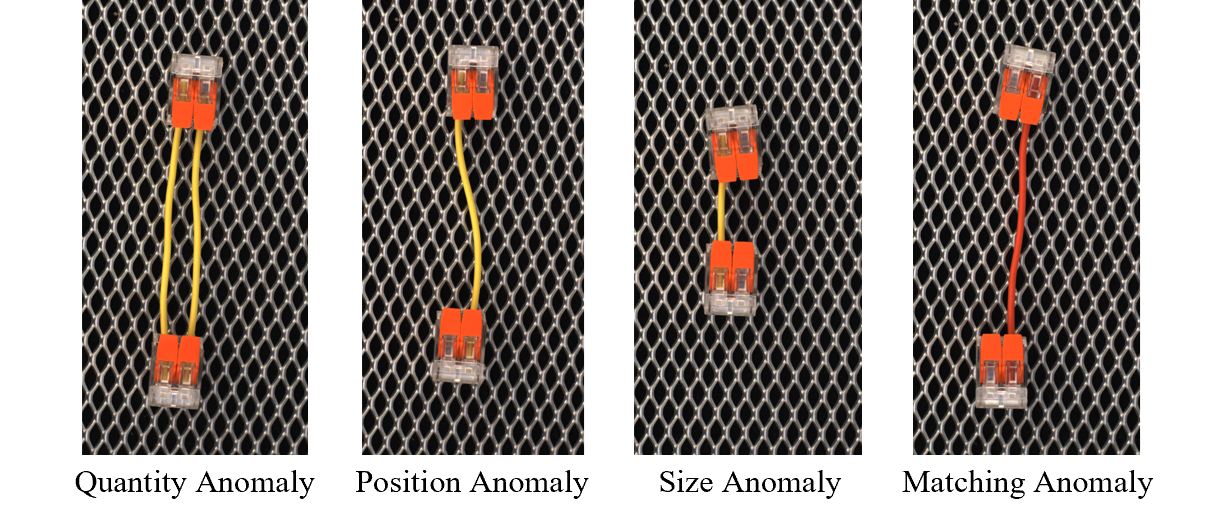}
\caption{\textbf{Quantity Anomaly:} The expected count for the cable is one, yet two is present. \textbf{Position Anomaly:} Cable should attach to the connectors with the same order defined by position; however, misplaced attachment is observed here.  \textbf{Size Anomaly:} The cable length is anticipated to fall within a specific range; however, it is observed to deviate from the normal range. \textbf{Matching Anomaly:} The cable color is expected to match connector number; however, a mismatch is observed here.}
\label{fig:anomalytype}
\end{figure}

LOCO-Annotations, comprising 2908 meticulously annotated images (1772 training and 1136 testing images, spanning various categories such as breakfast boxes, screw bags, pushpins, splicing connectors, and juice bottles.), diverges from MVTec LOCO’s approach by solely concentrating on logical anomalies. It categorizes the logical anomalies into four main types based on scenarios: Quantity Anomalies, Size Anomalies, Position Anomalies, and Matching Anomalies, as demonstrated in Fig.~\ref{fig:anomalytype}. This focus stems from a need to delve deeper into high-level semantic inconsistencies often overlooked in existing logical anomaly detection methods. Unlike MVTec LOCO, which includes both logical and structural anomalies in its testing set, LOCO-Annotations dataset exclusively annotates logical anomalies in its testing subset. Each image in the LOCO-Annotations is accompanied by detailed JSON files providing pixel-level object segmentation and ground truth annotations formatted as “Anomaly Type: Specific Reason”, clarifying the reasoning behind each identified logical anomaly. Examples of MVTec LOCO and LOCO-Annotations are illustrated in Fig.~\ref{fig:LOCO-Annotations}. This level of detail is pivotal in training and evaluating logical anomaly detection methods. The unique contribution of LOCO-Annotations lies in its detailed focus on logical anomalies and their underlying reasons, providing essential data for testing and refining models like LogiCode and other LLM-driven methods. This approach aligns with the evolving needs of industrial quality control, where understanding the “why” behind logical anomalies is as crucial as detecting them. Additionally, its provision of data for decoupling individual objects within images further enhances the dataset's utility, enabling a more detailed and nuanced analysis that is critical for understanding the complex dynamics of logical anomalies in industrial settings.

By offering a dataset that emphasizes logic and context, we aim to change the paradigm in anomaly detection research, focusing on logical semantic inconsistencies and their implications in industrial settings. The LOCO-Annotations not only fills the gap in industrial anomaly detection research but also paves the way for future advancements, setting new standards for the development of LLMs-based inspection systems.

\subsection{LogiBench}
LogiBench is meticulously designed for evaluating the LogiCode framework, with an emphasis on the thorough analyses of logical anomaly detection in industrial scenarios. The benchmark’s construction is driven by the need to evaluate the framework's binary classification accuracy, code generation success rate, and reasoning accuracy: 

\textbf{Binary Classification Accuracy:} This metric assesses the performance of the LLM-generated codes in correctly identifying the presence of logical anomalies in images.

It involves comparing the outcomes of the model with the ground truth (Normal/Abnormal) provided in the LOCO-Annotations to calculate crucial binary classification metrics like accuracy and recall rate. It is evaluated based on accuracy, precision, recall and F1-score. 

\textbf{Code Generation Success Rate:} This metric measures the LLM’s ability to generate executable Python codes based on the provided logical rules and request prompts. 

This metric specifically focuses on comparing the explanations generated by LLMs with the standard truth reasons provided in the LOCO-Annotations, thereby determining the correctness of the LLM's reasoning against established benchmarks. It is crucial due to the potential variability in LLMs' ability to generate Python codes for logical anomaly detection.

The success rate is determined by analyzing whether the generated codes run without syntax errors and correctly implement visual APIs.

\textbf{Reasoning Accuracy:} This metric evaluates the precision of LLM-generated codes in explaining the reasons for logical anomalies. It is assessed to address the observed instances where LLMs correctly classify anomalies but may provide inaccurate explanations for their reasoning. The assessment is two-fold, detailed as follows:

Human Evaluation: An expert analysis to validate the consistency and correctness of the explanations provided by the LLM. It is necessary due to the complexity involved in discerning anomaly reasons, requiring expert knowledge to validate the consistency and correctness of the LLM's explanations.

\begin{figure}[t]
\centering\includegraphics[keepaspectratio]{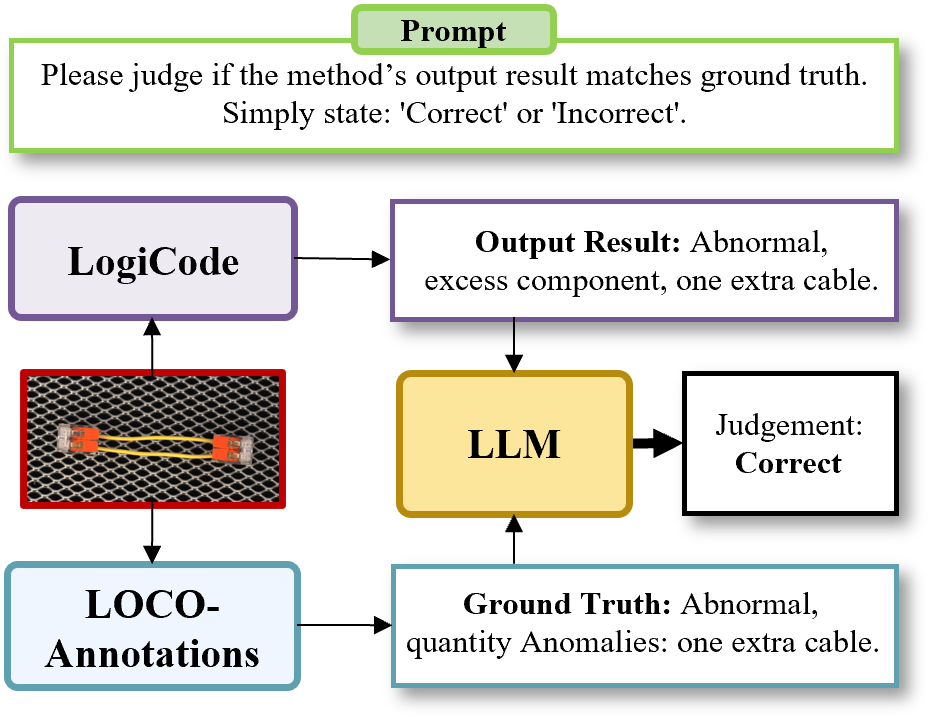}
\caption{Illustration for LLM Automatic Evaluation.}
\label{fig:llm_eval}
\end{figure}

LLM Automatic Evaluation~\cite{wang2023mint},~\cite{bi2023oceangpt}: Utilizes a structured prompt to compare the output result against the ground truth reason, evaluating the LLM’s accuracy in identifying logical anomaly reasons, as shown in Fig.~\ref{fig:llm_eval}. It is included to reduce the workload of manual evaluation.

LogiBench sets a new precedent for evaluating logical anomaly detection algorithms, providing a detailed and multidimensional evaluation of the LogiCode framework’s capabilities. Through this benchmarking process, we aim to demonstrate the advanced interpretability and efficiency of LogiCode in industrial scenarios, underlining its adaptability and robustness in a variety of industrial contexts. The thorough assessment provided by LogiBench also serves as a valuable resource for future enhancements and applications of LLMs in logical anomaly detection.

\subsection{Evaluation Results and Analyses}
The effectiveness of LogiCode, as assessed by the LogiBench benchmark, is presented in this subsection. Our comprehensive evaluation covers several key performance metrics, the results of which are summarized below.

At the outset, it is important to note that all experiments conducted in this study are based on GPT-4~\cite{openai2023gpt}. Employing GPT-4 for this purpose reflects the framework's integration of state-of-the-art LLM capabilities to ensure the precision of its anomaly reasoning analysis. The outcomes of the various performance metrics assessed are presented below.

\textbf{Binary Classification Accuracy:} The accuracy, precision, recall, and F1-score of the framework in identifying logical anomalies are measured against the ground truth labels provided by the LOCO-Annotations. 

\begin{table}[t]
\centering
\caption{Performance of the proposed method}
\label{tab:table1}
\resizebox{1.0\linewidth}{!}{
\begin{tabular}{@{}lccccc@{}}
\toprule[1.5pt]
Category              & Accuracy & Precision & Recall & F1 Score  \\ \midrule
Juice   bottle        & 0.992    & 0.987     & 1.000  & 0.994     \\
Breakfast   box       & 0.989    & 0.976     & 1.000  & 0.988     \\
Pushpins              & 0.990    & 0.976     & 1.000  & 0.988    \\
Screw bag           & 0.981    & 0.965     & 1.000  & 0.982     \\
Splicing connectors & 0.989    & 0.978     & 1.000  & 0.989     \\ \midrule
Average               & 0.989    & 0.976     & 1.000  & 0.988    \\ \bottomrule[1.5pt]
\end{tabular}
}
\end{table}

\begin{figure*}[ht!]
\centering\includegraphics[width=1.\linewidth]{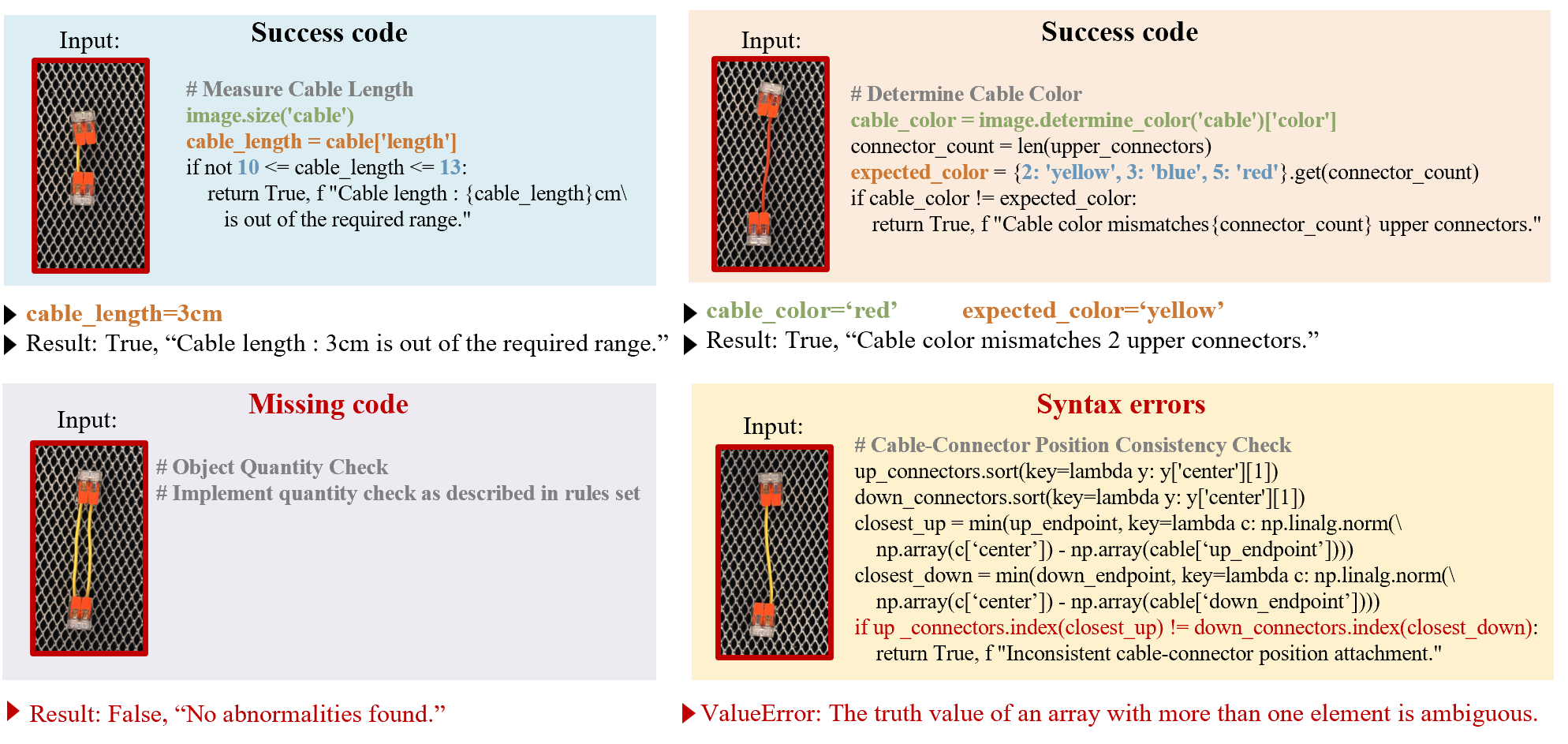}
\caption{Demonstration for success code, missing code and syntax error cases. The figure shows the core part of the generated code, the value of intermediate variables during the execution and the final output.}
\label{fig:code_exam}
\end{figure*}

The binary classification metrics are determined through comprehensive testing across all categories. The evaluation involved conducting five sets of experiments for each category, with the results averaged to obtain a reliable measure of accuracy, precision, recall and F1-score. This approach ensures a thorough and unbiased assessment of the framework's ability to detect logical anomalies.

The results for the Binary Classification metrics are presented in Table~\ref{tab:table1}. It can be observed that the binary classification metrics for anomaly detection are remarkably high, suggesting that the proposed framework can effectively detect logical anomalies.

\begin{table*}[h]
\centering
\caption{Accuracy comparison with the sota methods}
\label{tab:compare}
\begin{tabularx}{\textwidth}{@{}l *{5}{>{\centering\arraybackslash\hsize=.9\hsize}X} *{2}{>{\centering\arraybackslash\hsize=1.1\hsize}X} ccccc @{}}
\toprule[1.5pt]
Category/Methods & ST & PEFM & RD4AD & SPADE & GCLF & EfficientAD-S & EfficientAD-M & \textbf{LogiCode} \\ \midrule
Juice bottle & 0.741 & 0.731 & 0.767 & 0.806 & 0.935 & 0.970 & 0.976 & \textbf{0.992} \\
Breakfast box & 0.746 & 0.795 & 0.735 & 0.695 & 0.693 & 0.753 & 0.778 & \textbf{0.989} \\
Pushpins & 0.668 & 0.681 & 0.699 & 0.568 & 0.823 & 0.916 & 0.897 & \textbf{0.990} \\
Screw bag & 0.537 & 0.571 & 0.579 & 0.642 & 0.627 & 0.657 & 0.672 & \textbf{0.981} \\
Splicing connectors & 0.634 & 0.639 & 0.621 & 0.708 & 0.866 & 0.907 & 0.936 & \textbf{0.989} \\ \midrule
Average & 0.665 & 0.683 & 0.680 & 0.684 & 0.789 & 0.841 & 0.852 & \textbf{0.989} \\ \bottomrule[1.5pt]
\end{tabularx}
\end{table*}

To comprehensively evaluate the efficacy of the proposed LogiCode framework in detecting logical anomalies, comparative analyses against current state-of-the-art (SOTA) methods have been conducted. As outlined in Table ~\ref{tab:compare}, our assessment is meticulously focused on logical anomalies across various categories, contrasting our binary classification accuracy with that of established approaches. Unlike traditional methods, which predominantly rely on AUROC metrics for performance evaluation, the proposed framework necessitates a binary classification due to its inherent design to output discrete labels (0 for 'normal' and 1 for 'anomaly'). This differentiation necessitated the adoption of an alternative evaluation metric, focusing on maximized accuracy through variable threshold optimization.

The implementation of LogiCode, along with all competing methods, has been conducted in-house to ensure a fair and controlled comparison. Notably, the use of the LOCO-Annotations dataset allows us to leverage additional data, not merely for performance enhancement but as a means to showcase the inherent superiority of the proposed framework. The results, as depicted in Table ~\ref{tab:compare}, affirm the exceptional performance of the proposed method, underscoring its potential to redefine standards in logical anomaly detection within industrial settings.

\textbf{Code Generation Success Rate:} This paper evaluated the success rate of the framework in generating syntactically correct and executable Python codes. 

To evaluate the success rate of code generation, code produced for each category are generated 20 times separately. The success rate is calculated based on the number of times the codes ran successfully and aligned with the logical anomaly rules without any Python syntax errors. This repeated execution method provides a robust assessment of the framework's consistency in generating executable Python codes.


\begin{table}[h]
\centering
\caption{Code Generation Success Rate}
\resizebox{1.0\linewidth}{!}{
\label{tab:table2}
\begin{tabular}{p{1.cm}<{\centering} p{1.0cm}<{\centering} p{1.0cm}<{\centering} p{1.0cm}<{\centering} p{1.0cm}<{\centering} p{1.4cm}<{\centering} p{1.0cm}<{\centering}}
\toprule[1.5pt]
\multirow{2}{*}{Metric}     & \multirow{2}{*}{\makecell{Juice \\bottle}} & \multirow{2}{*}{\makecell{Breakfast \\box}} & \multirow{2}{*}{Pushpins} & \multirow{2}{*}{\makecell{Screw \\bag}} & \multirow{2}{*}{\makecell{Splicing \\connectors}}  & \multirow{2}{*}{Average}  \\ 
& & & & & & \\ \midrule
Success & 0.350        & 0.650         & 0.800    & 0.600     & 0.550               & 0.590   \\
Error   & 0.300        & 0.100         & 0.100    & 0.500     & 0.150               & 0.140   \\
Missing & 0.350        & 0.250         & 0.100    & 0.350     & 0.300               & 0.270   \\ \bottomrule[1.5pt]
\end{tabular}
}
\end{table}


Table~\ref{tab:table2} presents the success rate of code generation using prompts in GPT-4, where “success” is attributed to instances where the codes’ functionality aligns with the logical anomaly rules without Python syntax errors that could prevent execution. The other instances are categorized as “missing” or “error,” as detailed in the Table~\ref{tab:table2}. The examples for the generated codes are shown in Fig.~\ref{fig:code_exam}.

It is observed that due to the current instability of LLMs, the code generation success rate presently hovers around 60\%. However, it is expected that future iterations of LLM frameworks will significantly improve this metric.

\textbf{Reasoning Accuracy:} The framework’s precision in articulating the reasons for anomalies has been tested through LLM automatic evaluation and human evaluation. 

For reasoning accuracy, the evaluation focused on codes that successfully executed in each category. Five sets of such successful codes are analyzed for each category, with the results averaged to assess the framework's precision in articulating the reasons for anomalies. 


\begin{table}[h]
\centering
\caption{Human and LLM evaluation Reasoning Accuracy}
\label{tab:table3}
\resizebox{1.0\linewidth}{!}{
\begin{tabular}{p{0.9cm}<{\centering} p{1.0cm}<{\centering} p{1.0cm}<{\centering} p{1.0cm}<{\centering} p{1.0cm}<{\centering} p{1.4cm}<{\centering} p{1.0cm}<{\centering}}
\toprule[1.5pt]
\multirow{2}{*}{Methods}     & \multirow{2}{*}{\makecell{Juice \\bottle}} & \multirow{2}{*}{\makecell{Breakfast \\box}} & \multirow{2}{*}{Pushpins} & \multirow{2}{*}{\makecell{Screw \\bag}} & \multirow{2}{*}{\makecell{Splicing \\connectors}}  & \multirow{2}{*}{Average}  \\ 
& & & & & & \\
\midrule
Human & 0.901        & 0.974         & 0.976    & 0.988     & 0.976               & 0.963   \\
LLM   & 0.869        & 0.928         & 0.932    & 0.936     & 0.954               & 0.924   \\ \bottomrule[1.5pt]
\end{tabular}
}
\end{table}


\textit{Human Evaluation:} The human evaluation involved employing three researchers with relevant backgrounds to assess the framework's reasoning accuracy. Each researcher independently evaluates the explanations provided by the framework, and their assessments are averaged to determine the overall accuracy. 

\textit{LLM Evaluation:} For the automatic evaluation of reasoning accuracy, GPT-4 is employed. The implementation details of this evaluation are provided in the LogiBench benchmark section.

As shown in Table~\ref{tab:table3}, the comparison between the  automatic reasoning evaluation of LLM and the assessments by human experts indicates a close correspondence, suggesting the reliability of the LLM’s self-judgment in identifying the causes of logical anomalies. Moreover, the reasoning accuracy metric for the framework consistently exceeds 90\%, affirming the framework’s efficacy.

These results provide crucial insights into the capabilities and potential areas of improvements for the LogiCode framework in the context of logical anomaly detection in industrial settings.

\subsection{Discussions}
The LogiCode framework introduces advancements and addresses limitations in existing LLMs in the context of industrial logical anomaly detection. This subsection explores the impact of LogiCode on this field, emphasizing both its advantages and the constraints of current LLMs.

\noindent \textbf{Impact on Industrial Logical Anomaly Detection:} LogiCode, integrating LLMs and specialized APIs, enhances the accuracy of detecting high-level semantic inconsistencies. This is particularly crucial in industrial settings where such anomalies can significantly affect product quality and safety. By providing more accurate and interpretable anomaly detection, LogiCode has the potential to transform quality control processes, reducing both time and costs associated with manual inspections.

\noindent \textbf{Advantages and Limitations of LLM Methods:} LLMs contribute a nuanced understanding of complex logical relationships, generating contextually relevant codes for anomaly detection—a significant advancement from traditional rule-based systems. Despite their capabilities, LLMs may encounter challenges in highly specialized or niche industrial scenarios with limited training data. Furthermore, the black-box nature of these models can pose transparency and trustworthiness issues in critical applications. 

\noindent \textbf{Future Prospects and Recommendations: } Future research could explore the integration of reinforcement learning and LLMs to autonomously summarize logical rules, reducing dependence on expert inputs and enhancing adaptability to new scenarios. The current reliance on detailed annotations for API functions such as “find” could evolve into more sophisticated zero-shot or one-shot methods, enabling generalized object segmentation without heavy reliance on pixel-level labels. This evolution would broaden the framework’s applicability across diverse industrial settings and decrease the need for extensive dataset preparations.

\section{Conclusions}
This paper introduces the LogiCode framework, a novel approach that leverages LLMs for industrial logical anomaly detection. LogiCode represents a significant transition from traditional methods, focusing on high-level semantic inconsistencies and offering a more nuanced understanding of logical anomalies. Its integration of LLMs for code generation and logical reasoning sets a new benchmark in the field, demonstrating remarkable adaptability and interpretability across various industrial scenarios.

Through rigorous evaluations with the LogiBench benchmark, LogiCode shows superior performance in binary classification accuracy, code generation success rate, and reasoning accuracy. These achievements underscore the framework’s efficiency and reliability in anomaly detection and its potential in revolutionizing quality control processes in industrial settings.

Looking forward, the potential of LLM-driven anomaly detection is vast. Future work will focus on further enhancing the models’ autonomy, allowing them to adapt seamlessly to a variety of scenarios with minimal human intervention. Additionally, efforts to improve the generalizability of the models to maintain consistent performance across different industries will be crucial.

In summary, LogiCode is more than just a solution to current challenges in logical anomaly detection, it is a foundation for a future where intelligent models are integral to industrial processes, offering robust, interpretable, and efficient solutions. The ongoing developments in LLMs and their applications in this field open up exciting possibilities for industrial anomaly detection, leading the way toward more advanced, reliable, and automated quality control systems.

\ifCLASSOPTIONcaptionsoff
  \newpage
\fi



%

{\small
\bibliographystyle{unsrt}

\bibliography{ref}
}
%
\begin{IEEEbiography}[{\includegraphics[width=1in,height=1.25in,clip,keepaspectratio]{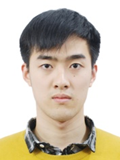}}]{Yiheng Zhang}
received the B.S. degree from the School of Mechanical Science and Engineering, Huazhong University of Science and Technology, Wuhan, China, in 2022, where he is currently pursuing the M.S. degree in mechanical engineering. His current research interests include industrial foundation models and their real-world applications, visual understanding, anomaly detection, and computer vision.
\end{IEEEbiography}

\begin{IEEEbiography}[{\includegraphics[width=1in,height=1.25in,clip,keepaspectratio]{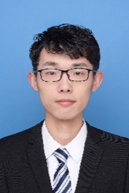}}]{Yunkang Cao}
(Student Member, IEEE) received the B.S. degree in mechanical design, manufacturing and automation from Huazhong University of Science and Technology (HUST), Wuhan, China, in 2020, where he is currently pursuing the Ph.D. degree in mechanical engineering. 
His current research interests include industrial foundation models and their real-world applications, anomaly detection, and computer vision.

\end{IEEEbiography}

\begin{IEEEbiography}[{\includegraphics[width=1in,height=1.25in,clip,keepaspectratio]{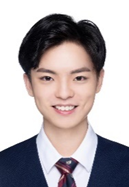}}]{Xiaohao Xu}
received his B.S. degree in mechanical design, manufacturing and automation from Huazhong University of Science and Technology (HUST), Wuhan, China in 2022. He is currently pursuing the Ph.D. degree in robotics with the University of Michigan, Ann Arbor, MI, USA.
His current research interests include the fundamental theory and real-world applications of robotics, computer vision, and video understanding.

\end{IEEEbiography}

\begin{IEEEbiography}[{\includegraphics[width=1in,height=1.25in,clip,keepaspectratio]{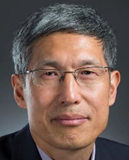}}]{Weiming Shen}
 (Fellow, IEEE) received the B.E. and M.S. degrees in mechanical engineering from Northern Jiaotong University, Beijing, China, in 1983 and 1986, respectively, and the Ph.D. degree in system control from the University of Technology of Compiegne, Compiegne, France, in 1996. He is currently a Professor with the Huazhong University of Science and Technology (HUST), Wuhan, China, and an Adjunct Professor with the University of Western Ontario, London, ON, Canada. Before joining HUST in 2019, he was a Principal Research Officer at the National Research Council Canada. He is a Fellow of Canadian Academy of Engineering and the Engineering Institute of Canada. 
His work has been cited more than 20000 times with an h-index of 70. He authored or coauthored several books and more than 560 articles in scientific journals and international conferences in related areas. His research interests include agent-based collaboration technologies and applications, collaborative intelligent manufacturing, the Internet of Things, and Big Data analytics

\end{IEEEbiography}

\end{document}